\documentclass[10pt,twocolumn,letterpaper]{article}

\usepackage{cvpr}
\usepackage{times}
\usepackage{epsfig}
\usepackage{graphicx}
\usepackage{amsmath}
\usepackage{amssymb}
\usepackage{textcomp} 
\usepackage{mathtools}
\usepackage{gensymb} 
\usepackage{xfrac} 
\everymath{\displaystyle} 
\usepackage{siunitx} 
\usepackage{booktabs}
\usepackage{tabularx}
\usepackage{caption} 
\usepackage{diagbox}

\newcommand\ytrue{y} 
\newcommand\ypred{\hat{y}} 

\usepackage[breaklinks=true,bookmarks=false]{hyperref}
\usepackage[square,comma,numbers,sort&compress,sectionbib]{natbib}

\cvprfinalcopy 

\def\cvprPaperID{2} 

\ifcvprfinal\fi
\begin{document}

\title{On indirect assessment of heart rate in video}

\author{Mikhail Kopeliovich  \hspace{35pt} Konstantin Kalinin \hspace{35pt} Yuriy Mironenko  \hspace{35pt} Mikhail Petrushan\\
Southern Federal University, Center of Neurotechnologies\\
Rostov-on-Don, Russian Federation\\
{\tt\small kop@km.ru}
}

\maketitle

\begin{abstract}
Problem of indirect assessment of heart rate in video is addressed. Several methods of indirect evaluations~(adaptive baselines) were examined on Remote Physiological Signal Sensing challenge. Particularly, regression models of dependency of heart rate on estimated age and motion intensity were obtained on challenge\textquotesingle s train set. Accounting both motion and age in regression model led to top-quarter position in the leaderboard. Practical value of such adaptive baseline approaches is discussed. Although such approaches are considered as non-applicable in medicine, they are valuable as baseline for the photoplethysmography problem.
\end{abstract}

\section{Introduction and prior work}
RePSS (The 1st challenge on Remote Physiological Signal Sensing) was held in conjunction with CVPR~2020~\cite{Li2020, Li2018, Yu2019, Niu2019}. The challenge aimed at comparing accuracy of different methods of heart rate assessment in video with stationary and moving humans. Common approaches on heart rate estimation in video are mostly based on remote photoplethysmography (rPPG) which is analysis of skin color changes caused by heart beat~\cite{Verkruysse2008, mcduff2019} or ballistocardiography which is detection of tiny motions caused by heart pulsation~\cite{Balakrishnan2013}.

We, however, tried to indirectly ``guess'' a heart rate. Our guess was based on assessment of age and motion in the video, without using of any PPG-related signal at all. We called this approach ``adaptive baseline''.

The inverse problem was, to great extent, solved. Heart rate (HR) as a proxy variable for evaluation of energy expenditure is a well-known and widely used technique~\cite{HRProxy1992, HRProxy2012}. This technique typically involves the use of other factors, such as body weight index~(BWI), gender or age~\cite{HRProxy2005, HRProxyWithoutCalibration2001}, to produce more accurate estimation. As long as such estimation is accurate enough for the practical usage~\cite{FirstbeatEnergyExpenditure}, the following question appears: is it possible to invert this scheme to evaluate HR itself? Is it possible to combine the information about BWI, age, gender, physical activity or other factors, and produce good HR guess?

It turns out that our adaptive baseline was actually better than $3/4$ of other participants\textquotesingle best efforts. It is a matter of discussion whether the baseline HR estimations are of any practical value, even if formally accurate enough.
\section{The challenge description}
The RePSS dataset is divided into training and testing sets. The training set consists of randomly cut 10-second clips from 500 randomly selected videos of VIPL-HR~V2 dataset~\cite{Li2020}. Up to date, this dataset is poorly described by the authors. Regarding the training set, it contains 500~folders of 10~sec  videos and metadata. In most cases, there are five videos per folder, cut from corresponding video files of the VIPL-HR~V2 dataset, resulting in 2498~videos. Metadata of a single ground truth HR value and average value of frame rate correspond to each video.
Authors state that they used a Realsense~F200 camera to record videos with 960$\times$720 resolution~\cite{Li2020}. However, the resolutions of the training set videos vary.
The scene backgrounds are non-monotonous and sometimes dynamic with presence of other people behind the subject.
Subjects in the dataset are of different age (including children and elderly) and gender. Total number of subjects is 500.

The testing set contains 1000 video files of 10~sec videos from 200 subjects (100 from VIPL-HR~V2 and 100 from OBF Database~\cite{Li2018}). The extra complexity is that the areas of the eyes and mouth areas are covered with mosaics. Instead, the authors provided the facial bounding box location and 68 facial landmarks.

The RePSS authors selected the mean average error~(MAE) as primary metric to compare methods of HR evaluation. Two additional metrics were used: root mean square error~(RMSE) and Pearson correlation coefficient~(R).
\section{Methods}
In rPPG, average HR value can be used as baseline. It is essentially a prior knowledge about HR distribution on a training set for a proposing rPPG method. Average HR is the simplest form of such prior knowledge. More sophisticated forms may involve additional information such as estimated age and gender of a person in video, type of activity, and others~\cite{Quer2020}. These additional factors can adapt baseline to particular conditions observable in video under analysis. In particular, we studied relations of HR on estimated age of a person and on person\textquotesingle s motion intensity in challenge’s training dataset. Further, we estimated age and motion activity in videos from test set and applied prior knowledge on HR dependencies to obtain baseline predictions.

We consider four methods for heart rate estimation: Baseline by Constant HR~(BC), Baseline by Motion~(BMotion), Baseline by Age~(BAge), and Baseline by Age and Motion~(BAM).
\paragraph{BC}
represents a constant function returning the same HR value $C$ for any input.
According to~\cite{Quer2020}, normal heart rate for healthy adult at rest could be estimated as $C=70$~bpm. However, regarding RePSS, the average HR on the training set was 87.95~bpm. We rounded this value down ($C=87$~bpm) and used it as BC output. The rounding down was made both for sake of simplicity and basing on testing set analysis: it is half composed of videos from OBF Database, where, for some sessions, subjects are expected to sit still before the physical exercises~\cite{Li2018}. Due to relatively high HR on the training set, we assume that average HR on testing set would be slightly lower because of before-exercise part of the OBF.
\paragraph{BMotion}
is based on HR prediction by motion amplitude. Unlike ballistocardiography~\cite{Balakrishnan2013}, BMotion method estimates HR by an average motion amplitude~(AMA), which is a single value generated per video: 
\begin{equation}
\text{AMA} = \frac{1}{F-\delta}\sum_{f=\delta+1}^F
    \left|
    \frac{y_c^f}{H_\text{face}^f} - \frac{y_c^{f-\delta}}{H_\text{face}^{f-\delta}}
    \right|
,
\end{equation}
where $f$ is a frame index, $F$ is a total frame count, $y_c^f$ is a vertical coordinate of facial rectangle in pixels, $H_\text{face}^f$ is a height of facial rectangle in pixels, $\delta$~(=10) is an offset to calculate motion amplitude. The motion amplitude is estimated in each frame as displacement of relative vertical coordinate of facial rectangle after $\delta$ frames.
We didn\textquotesingle t exploit horizontal coordinate of facial rectangle due to noisy value of width of the detected rectangle caused by head rotations.
To detect facial rectangle, we used Multitask Cascaded Convolutional Networks~\cite{Zhang2016_2} implemented on Python by Centeno~\cite{Centeno2019}.

Once calculated, the AMA values and corresponding actual HR values are used as inputs for linear regression algorithm. The resulting coefficients of approximation line are used to generate HR predictions.
\paragraph{BAge}
is based on age estimation.
Age was estimated on cropped facial image as one of 26 classes corresponding to a range of 15 to 40~years by Consistent Rank Logits framework~\cite{Cao2019} implemented on Python by~\cite{mshehrozsajjad2019}. Age values were calculated for 10 evenly selected frames in video; the overall age was estimated by averaging their 5 median values.

As for BMotion, we used linear regression to generate HR predictions.
\paragraph{BAM}
combines the previous two approaches. First, BAge is used to train the basic model of linear regression providing array of $\mathbf{\ypred}^{\text{BAge}}$ estimations on the training set. Second, the relative errors
${\left(\ytrue_i - \ypred_i^{\text{BAge}}\right)}
\Big/
{\ypred_i^{\text{BAge}}}$
are used as ground truth data to train another linear regression model with the corresponding AMA values on input. Both the models applied similarly on the testing set, then the outputs of the second model are back-scaled to heart rate values: 
$
\ypred_i^{\text{BAge}} + 
\ypred_i^{\text{BAge}} \ypred_i^{\text{BAM}}
$,
where $\mathbf{\ypred}^{\text{BAM}}$ are predictions of the second model.
\section{Results and Discussion}
\begin{table}[]
\centering
\begin{tabular}{@{}l|lll@{}}
\multicolumn{1}{c|}{\begin{tabular}[c]{@{}c@{}}\# or\\   method\end{tabular}} & \multicolumn{1}{c}{\begin{tabular}[c]{@{}c@{}}MAE,\\ bpm\end{tabular}} & \multicolumn{1}{c}{\begin{tabular}[c]{@{}c@{}}RMSE,\\ bpm\end{tabular}} & \multicolumn{1}{c}{R} \\ \midrule
1 & 6.94 & 10.68 & 0.75 \\
2 & 7.92 & 14.38 & 0.59 \\
3 & 8.95 & 14.16 & 0.54 \\
4 & 12.39 & 16.09 & 0.23 \\
5 & 12.46 & 16.20 & 0.19 \\
\textbf{BAM} & \textbf{12.49} & \textbf{15.84} & \textbf{0.19} \\
\textbf{BAge} & \textbf{12.51} & \textbf{15.85} & \textbf{0.18} \\
7 & 12.54 & 16.09 & 0.10 \\
8 & 12.55 & 16.06 & 0.10 \\
9 & 12.58 & 15.97 & 0.10 \\
\textbf{BMotion} & \textbf{12.68} & \textbf{15.93} & \textbf{0.10} \\
10 & 12.74 & 16.13 & 0.10 \\
11 & 12.74 & 16.35 & 0.22 \\
12 & 12.91 & 16.54 & 0.08 \\
13 & 12.91 & 16.57 & 0.15 \\
14 & 12.91 & 16.57 & 0.15 \\
15 & 13.02 & 17.30 & 0.07 \\
\textbf{BC} & \textbf{13.26} & \textbf{16.29} & \textbf{0.00} \\
16 & 13.26 & 16.29 & 0.00 \\
17 & 13.29 & 16.75 & 0.02 \\
18 & 13.40 & 17.87 & 0.28 \\
19 & 13.55 & 17.46 & 0.16 \\
20 & 13.63 & 16.65 & 0.06 \\
21 & 14.37 & 18.75 & 0.11 \\
22 & 14.51 & 18.97 & 0.11 \\
23 & 14.76 & 19.11 & 0.04 \\
24 & 15.69 & 19.74 & 0.01 \\
25 & 20.12 & 25.55 & -0.02
\end{tabular}
\caption{Leaderboard of the RePSS competition. Results are sorted by MAE, which was a primary metric of the challenge. Proposed baselines are in bold.}
\label{tab:leader}
\end{table}
RePSS challenge was one of first attempts to evaluate and compare different approaches of remote HR estimation on moving person in standardized conditions. The Table~\ref{tab:leader} provides the challenge leaderboard along with our proposed baselines.

BC is the simplest method for indirect heart rate assessment. Surprisingly, it provided lower MAE and RMSE values than one third of the all other methods. Due to zero variance of BC\textquotesingle s answers its R value is undefined (set to zero). In order to calculate R value, one can add a slight random number to each answer of BC; this procedure wouldn\textquotesingle t significantly change MAE or RMSE.

The adaptive baselines, BMotion, BAge and BAM, further improved the result leading to the 6th place. The error of BAM is close to 4th and 5th places (less than 1\% difference).

Top three methods demonstrated MAE of $\approx$7-9~bpm, while the error of most methods was above 12~bpm. They are not far beyond even the simple average BC method with MAE of $\approx$13.25~bpm. Adaptive Baseline approach decreases MAE to $\approx$12.48~bpm.

\subsection{Possible Enhancements of the Baseline Method}
Even most sophisticated of our baselines, BAM, only uses age and motion assessments. Moreover, it uses very simple form of motion assessment and combines estimations in the very simple, linear way. Considering additional factors and more sophisticated combination of them would lead to even better estimations.

Obvious additional factors to consider is gender and body weight index (BWI), as long as they often used in HR-based techniques of indirect energy expenditure measurements. Some better form of the motion assessment should be considered, preferably one which correlates with energy expenditure. While BWI can hardly be determined from the facial video of the person, some signs of obesity or underweight still may be extracted.

Other source of inspiration is HR-based methods of indirect measurement of oxygen consumption~\cite{IndirectOxyUptake1, FirstbeatEnergyExpenditure} and blood pressure~\cite{IndirectBloodPressure1, IndirectBloodPressure2}. According to the inverse problem approach, they could be useful for the HR estimation. We assume that oxygen consumption and blood pressure themselves could be assessed via analysis of the average skin tone.

Third group includes features like pose, facial expression, speaking and others -- factors with uncertainly related to the heart rate, which, however, may be easily extracted from the video using already existing techniques. Therefore, they may be easily, almost for granted, tested for the applicability.

Finally, obvious enhancements is to use some simple neural network to combine all the pre-extracted factors into HR prediction.

\subsection{Dataset Bias}
Some factors could be dataset-specific. Particularly, if average HR is significantly different for different datasets (for example, children dataset concatenated with the adults dataset), some integral characteristics of these subsets (like different backgrounds) could be recognized and, thus, HR prediction baseline could be adapted to the subset's average. Although such accounting of all mentioned integral characteristics will likely lead to better prediction of HR and higher positions in challenges leaderboards, it doesn’t look like a way to create practically-valuable rPPG method. Moreover, it even may be done unintentionally.

\subsection{Concerns}
If adaptive baseline methods could be capable to produce HR predictions close to real values basing on subject’s appearance parameters is a matter of further studies. However, if they do, they will likely fail to predict HR deviations from values typical for recognized subject’s appearance. This is due to these methods do not extract pulse signal or its derivatives from data and mostly exploit prior knowledge on HR distribution over subject’s appearance. Such unrecognizable deviations are of prime importance in medical applications, as they often indicate abnormal health state and necessity of medical assistance.  

End-to-end deep models become popular in rPPG. In order to determine practical value of such models they shouldn\textquotesingle t be considered as black boxes generating good or bad predictions of HR values. Analysis should be performed to determine what kind of features are inferring in end-to-end nets: are they appearance-related features or pulse signal-related features. If they are appearance features than method is doing adaptive baseline calculation which has minor value in medical tasks.

\subsection{Indirect assessment of Heart Rate Variability}
While we failed to find any attempts to solve the problem of indirect estimation for the HR itself, there are lot of researches of this kind for the heart rate variability (HRV). Influence of age, gender, BWI (especially in cases of obesity and underweight) and other factors on HRV is a field of ongoing studies~\cite{IndirectHRV1, IndirectHRV2, IndirectHRV3, IndirectHRV4, IndirectHRV5, IndirectHRV6, IndirectHRV7}. While no ready-made indirect assessment method was proposed yet, strong relations were found. Thus, the same indirect approach for measuring HRV can also be used. It will be susceptible to the same flaws.
\paragraph{Acknowledgments.} 
\ifcvprfinal 
The project is supported by the Russian Ministry of Science and Higher Education in the framework of Decree No.~218, project No.~2019-218-11-8185 ``Creating a software complex for human capital management based on neurotechnologies for enterprises of the high-tech sector of the Russian Federation'' (Internal number HD/19-22-NY).
\else
anonymized \\[6\baselineskip]
\fi
\section{Conclusion} 
The adaptive baseline for the HR assessment, based solely on the analysis of the subject appearance, is proposed and discussed. Even linear regression model of the dependence of heart rate on age and intensity of movements fell into the top 25\% of the RePSS competition leaderboard. We consider such indirect assessment as valuable baseline for photoplethysmography problem, while non-applicable for medical applications. We outlined our concern about possible implicit indirect HR assessment in current end-to-end deep networks, which merits further study.
{\small
\bibliographystyle{ieeetr}
\bibliography{paper}
}
\end{document}